# DIRECTED CYCLES IN BELIEF NETWORKS

*Wilson X. Wen*

Department of Computer Science,
The University of Melbourne,
Parkville, 3052, Australia.

## 1. Introduction

The most difficult task in probabilistic reasoning may be handling directed cycles in belief networks. To the best knowledge of this author, there is no serious discussion of this problem at all in the literature of probabilistic reasoning so far. Indeed, for most probabilistic reasoning methods, it is very difficult to obtain an initial distribution if there are any directed cycles in the network. To avoid an exponential explosion of the number of states as the number of variables in the space increases, many methods to decompose the underlying networks have been proposed. However, almost all of these methods are based on the assumption that there are no directed cycles in the underlying networks.

On the other hand, there are cases where reasoning in a belief network with directed cycles is inevitable. For example

(1) A domain expert may refuse to simplify a network with directed cycles (see, for example, Fig. 4.1.) to an acyclic one.

(2) probabilistic knowledge bases may have been extracted from some sample data [15] in which there are some cyclic conditional constraints on the underlying spaces.

In these cases, an attempt to oversimplify a complex network to an acyclic one may produce an incorrect or inaccurate result. Therefore, we have to develop some methods to handle cycles. Among the existing methods, there are methods which may overcome the difficulty. These are the methods based on information theory, in particular, the principles of Maximum Entropy (ME) and Minimum Cross Entropy (MCE). However, almost all of the authors who propose these methods never mention this feature of their methods and thus never explore an efficient way to make use of it.

In this paper, we discuss some problems of reasoning in belief networks with directed cycles. We argue that the initial distribution for the network can be easily obtained by ME/MCE methods although it is very difficult (if possible) to get this by other rule-based methods when the underlying network contains directed cycles. The issues of decomposition of the underlying networks with directed cycles and detection of inconsistency in the constraints are also discussed, based on information theory and the theory of Markov random fields and Gibbs fields.

## 2. An Overview

Two years ago, Ross Quinlan asked me how to obtain the distribution of a space of two variables $A$ and $B$ in Fig. 2.1 if we know the conditional probabilities $P(A|B)$ and $P(B|A)$ and whatever else may be consistent with the conditional probabilities. Indeed, even this very simple problem is enough to challenge most of the existing probabilistic reasoning methods. The only reason is that it is a directed cycle. However, as he observed, this directed cycle makes no trouble at all for ME/MCE methods. The only thing we need to do is to write the two conditional probabilities as two linear equality constraints and solve the ME/MCE problem by the traditional Lagrange multiplier method.

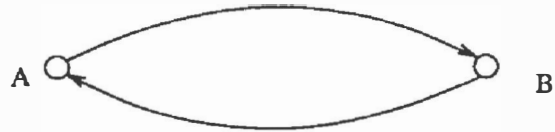

Fig. 2.1 A directed Cycle

Among the existing probabilistic reasoning methods, the following methods imply the above idea:

### 2.1. Lemmer's method

Lemmer [7] derives Jeffrey's rule [3] from the MCE principle. He also proposes a method to decompose the underlying networks into a tree of Local Event Groups (LEG's), and propagates beliefs around LEG's according to Jeffrey's rule. Although Lemmer's decomposition method only handles acyclic networks and with single root nodes, the concept of trees of LEG's itself does not imply any acyclic assumption. There may be some small directed cycles inside some LEG's. We can still use Jeffrey's rule to propagate the beliefs around the LEG's provided we have methods to decompose the underlying networks and to obtain a prior distribution for each LEG.

### 2.2. Cheeseman's Method

By maximizing the entropy of the underlying distribution subject to any constraints, Cheeseman's method [2] can always obtain the correct prior distributions even though there are some directed cycles in the underlying networks. To avoid an exponential explosion of the number of states as the number of variables in the networks increases, Cheeseman also proposes an efficient method to perform the relevant summations. Because this method needs to group the summations, it handles only some small directed cycles each of which can occur inside a single group.

### 2.3. Wise's Method

Neither Lemmer nor Cheeseman address explicitly the problem of handling directed cycles using their methods, although their methods have the ability to do so. The only author who addresses this problem explicitly is Wise. In Chapter 9 of his Ph. D. thesis [16], he wrote

> Unfortunately, attempts to produce this type of reasoning in a strictly rule-based system run into severe problems, not only in searching through an exponentially growing number of implicit backward or crossing arrows, but also because of cycles. For example, if $x$ supports $y$, then $y$ supports $x$, and a positive feedback cycle could drive both to probability 1. Also, if $\neg x$ supports $\neg y$, then $\neg y$ supports $\neg x$, and positive feedback could drive both to probability 0. ... To the ME/MXE method, explicit statement of such backward arrows simply provides redundant constraints, with absolutely no effect. As pointed out earlier, these cycles are no trouble at all (for) the ME/MXE method - so long as they are not inconsistent. This may be seen from the facts that data is input as linear constraints, not rules, ...

In his thesis, Wise only addresses directed cycles of length 2, but even this simplest case is enough to show what is the difference between ME/MCE methods and other methods when they are used to handle directed cycles.

### 2.4. Pearl's Graphical Criterion

Although Pearl's graphical criterion for testing conditional independence [9] was proposed for acyclic belief networks, we have generalized this criterion to the case with directed cycles based on information theory and the theory of Markov and Gibbs random fields [12].

### 2.5. Spiegelhalter's method

Similarly, Spiegelhalter's decomposition method [6] for belief networks can be also generalized in the case with directed cycles. Essentially, neither of the above two methods relies on the acyclic assumption. The main difficulty for Bayesian methods is to obtain a consistent initial distribution with a cyclic, and often incomplete, specification.

Although ME/MCE methods may be potential candidates for handling directed cycles in belief networks, there are still some problems to be solved before these methods can be used practically for this purpose. These problems are

(1) How to decompose the underlying networks to avoid an exponential explosion of the number of states as the number of variables in the network increases?

(2) After decomposition, How to obtain initial distributions of component marginal spaces which are consistent with each other globally?

Having decomposed the underlying network into small pieces and obtained an initial distribution, we can do the rest of the reasoning using some methods like [6,7,13]. In the following sections, we will try to answer these questions.

### 3. A Simple Example

To begin with, we give a simple example to show how to obtain an ME initial distribution for a cycle with two links $A \rightarrow B$ and $B \rightarrow A$ (Fig. 2.1).

Suppose we know $P(A|B)=0.7$ and $P(B|A)=0.8$. The ME distribution which satisfies these conditional constraints can be obtained in the following two ways [11]:

### 3.1. Numerical Method

According to [11], the above problem can be formulated as the following MCE problem:

$$\text{Minimize} \sum_{A,B} P(A,B) \log \frac{P(A,B)}{0.25}$$

$$\text{subject to} \sum_{A,B} P(A,B) = 1,$$

$$P(A|B) = 0.7, \quad P(B|A) = 0.8,$$

and the gradients of the Dual function are

$$D'_0 = 0, \quad D'_1 = 0.2, \quad D'_2 = 0.3.$$

Using the conjugate gradient method, we obtain the following MCE distribution:

$$P(\bar{A}, \bar{B}) = 0.2808, \quad P(\bar{A}, B) = 0.1836,$$
$$P(A, \bar{B}) = 0.1071, \quad P(A, B) = 0.4285.$$

### 3.2. Successive Updating

Suppose that a system $S$ of $m$ binary random variables $x_i$ ($0 \leq i < m$, m is a finite integer) has a set of $2^m$ possible micro-states $\{s_j \mid 0 \leq j < 2^m\}$ with distribution $p=\{P(s_j)\}$, and we have a prior distribution $p^{(0)}$ that estimates $p$. According to the values of $n$ distinct variables, $x_{i_k}$, ($0 \leq i_k < m$, $k=0, \ldots, n-1$, $n<m$), in $S$, we partition the state space $\{s_j\}$ into $2^n$ exclusive and exhaustive subspaces called macro-events $S_l$ ($0 \leq l < 2^n$), such that in each of these events the value of each $x_{i_k}$ is constant. We partition each $S_l$ further into two exclusive and exhaustive subspaces $S_{l_0}$ and $S_{l_1}$ according to another variable $x_{i_n}$ in $S$, ($x_{i_n} \notin \{x_{i_k}\}$), such that the value of $x_{i_n}$ is 0 and 1 in $S_{l_0}$ and $S_{l_1}$, respectively.

Suppose in addition to the prior of $S$ we also know a conditional constraint
$$\mu_l = P(x_{i_n} \mid x_{i_0}, \ldots, x_{i_{n-1}}) \quad (3.1)$$
It can be shown ([11], see also [16] for some more general cases) that the MCE posterior distribution is
$$\hat{P}(s_j) = \rho \, \sigma \, P^{(0)}(s_j), \quad s_j \in S_l, \quad (3.2)$$
where

(i) $\rho$ is a normalization factor to get a unit sum,

(ii) 
$$\sigma = \begin{cases} 1, & s_j \notin S_l, \\ \left( \dfrac{(1-\mu_l) \sum\limits_{s_{j'} \in S_{l_1}} P^{(0)}(s_{j'})}{\mu_l \sum\limits_{s_{j'} \in S_{l_0}} P^{(0)}(s_{j'})} \right)^\beta, & s_j \in S_l. \end{cases}$$

and
$$\beta = \begin{cases} \mu_l, & s_j \in S_{l_0}, \\ \mu_l - 1, & s_j \in S_{l_1}. \end{cases}$$

Starting from a uniform prior distribution (ie. we want an ME initial distribution) and using the above constraints $\mu_0=0.7$ and $\mu_1=0.8$ twice each, we obtain

$$P(\bar{A}, \bar{B}) = 0.2807, \quad P(\bar{A}, B) = 0.1808,$$
$$P(A, \bar{B}) = 0.1075, \quad P(A, B) = 0.4299.$$

It is easy to see that this result is only an approximation to the real MCE result which is obtained earlier by the numerical method. More iterative steps may be needed to get a more accurate approximation, although the above result is actually accurate enough for most practical applications.

It should be noted that not all combinations of conditional constraints are consistent. For example, it can be shown that the following combination
$$P(A \mid \bar{B}) = 0.2, \quad P(A \mid B) = 0.7,$$
$$P(B \mid \bar{A}) = 0.1, \quad P(B \mid A) = 0.8,$$
is not consistent. More generally, because all marginal constraints and conditional constraints are linear equality constraints, we have the following theorem:

**Theorem 3.1:** A set of linear equality constraints is consistent iff its corresponding set of linear equations has at least one non-zero solution.

For the above example, write the constraints in the format of linear equality:
$$\begin{cases} -0.2 \, P(\bar{A}, \bar{B}) + 0.8 \, P(A, \bar{B}) = 0 \\ -0.7 \, P(\bar{A}, B) + 0.3 \, P(A, B) = 0 \\ -0.1 \, P(\bar{A}, \bar{B}) + 0.9 \, P(\bar{A}, B) = 0 \\ -0.8 \, P(A, \bar{B}) + 0.2 \, P(A, B) = 0 \end{cases}$$

It is easy to see that this set of linear equations has vector $\mathbf{0}$ as its unique solution and therefor the constraint set is not consistent. Theoretically, Theorem 3.1 seems very simple, but the size of the set of equations will grow exponentially as the number of variables in the probabilistic space increases. Therefore, we need a method to check the consistency of the constraints locally rather than globally. We will discuss this issue later.

## 4. General Belief Networks and Their Markov Properties

Following the definitions in section 3.2, suppose we have the following constraint set $CS$ on the distribution $p$ of $S$, which may be elicited from domain experts or extracted from a sample database:

(1) *Conditional Constraints* ($CCS$): $\mu_l$ in (3.1).

(2) *Marginal Constraints* ($MCS$):
$$v_{l'} = P(x_{i'_0}, \ldots, x_{i'_{n'-1}})$$
For simplicity, we assume that these constraints can be added into $CS$ only when there are some corresponding $CCS$ $\mu_l \in CS$ such that
$$\{x_{i'_0}, \ldots, x_{i'_{n'-1}}\} \subseteq \{x_{i_0}, \ldots, x_{i_{n-1}}\},$$
although it is not necessary for our final conclusions.

(3) *Universal Constraints (UCS)*:
$$\sum_{j=0}^{2^m-1} P(s_j) = 1.$$

According to the CCS in CS, we may have the following definitions:

**Definition 4.1:** A *neighbor system* $\sigma$ in $S$ is a set of sets $\{\sigma x_i \mid x_i \in S, \sigma x_i \subset S\}$, such that

(i)  $x_i \notin \sigma x_i$,

(ii) $x_j \in \sigma x_i \leftrightarrow \exists \mu_l \in CS, x_i, x_j \in \{x_{i_0}, ..., x_{i_{n-1}}, x_{i_n}\}$.

Obviously, $x_i \in \sigma x_j \leftrightarrow x_j \in \sigma x_i$. Also, for $S' \subset S$,
$$\sigma S' = \{x_j \in S-S' \mid \exists x_i \in S', x_j \in \sigma x_i\}.$$

**Definition 4.2:** A *belief network* is a directed graph $G = <V, E>$, such that

(i)  The node set $V = S$,

(ii) The link set is
$E = \{<x_i, x_j> \mid \exists CCS \ \mu_l \in CS, j = i_n, i \in \{i_0, ..., i_{n-1}\}\}$

Given the above constraint set CS, we may calculate the ME distribution of $S$ using Lagrange multiplier method [2].

For any set $R$, we use $R+x, (x \notin R)$, and $R-x, (x \in R)$, as abbreviations of $R \cup \{x\}$ and $R-\{x\}$, respectively. If $R = \{x_{i_0}, ..., x_{i_{n-1}}\}$, we use $P(x_{i_n} \mid R)$ to stand for $P(x_{i_n} \mid x_{i_0}, ..., x_{i_{n-1}})$.

**Definition 4.3:** $S$ is a *Markov Random Field* (MRF) with respect to $\sigma$, if for any $S'$, $\sigma x_i \subset S' \subset S$ and $x_i \notin S'$, its distribution satisfies
$$P(x_i \mid S') = P(x_i \mid \sigma x_i).$$

In [12], we have proven the following theorem and corollary:

**Theorem 4.1:** With the ME distribution of $S$ subject to the constraint set $CS$, $S$ is an MRF with respect to the neighbor system $\sigma$.

**Definition 4.4:**

(i) A *potential* $U$ on $S$ is a family of functions
$$\{U_{S'}(x_0, ..., x_{m-1}) \mid S' \subset S\}$$
from $S$ to the real line, such that
$$U_{S'}(x_0, ..., x_{m-1}) = U_{S'}(x'_0, ..., x'_{m-1})$$
where $x_i = x'_i$ for all $x_i$ and $x'_i \in S'$ and $U_\varnothing = 0$.

(ii) The *energy* of $U$ is $H_U = \sum_{S' \subset S} U_{S'}$.

(iii) A set $C \subseteq S$ is called a *clique* if $\forall x_i, x_j \in C, i \neq j \rightarrow x_j \in \sigma x_i$. A clique $MC$ is called *maximal clique* if there is no other clique $C$, $MC \subset C$. Let $CC$ and $MCC$ be the classes of all cliques and maximal cliques in $S$, respectively. $U$ is called a *neighbor potential* if $\forall S' \notin CC, U_{S'} = 0$.

(iv) $S$ is a *Gibbs field* with potential $U$ if
$$P(x_0, ..., x_{m-1}) = \frac{1}{Z} e^{H_U(x_0, ..., x_{m-1})},$$
where $Z = \sum_{x_0, ..., x_{m-1}} e^{H_U(x_0, ..., x_{m-1})}$ is called the *potential function*. If $U$ is a neighbor potential, then $S$ is called a *neighbor Gibbs field* and $H_U = \sum_{C \in CC} U_C$.

**Corollary 4.1:** Under the condition of Theorem 4.1, $S$ is a neighbor Gibbs field with respect to $\sigma$.

According to Kemeny (p433, [4]), we have the following theorem:

**Theorem 4.2:** $S$ is an MRF iff
$P(S' \mid S'') = P(S' \mid \sigma S')$, where $\sigma S' \subset S'' \subseteq S$.

From Theorems 4.1 and 4.2, we may generalize Pearl's graphical criterion for testing conditional independence [9] to belief networks with directed cycles:

**Definition 4.5:**

(a) A subset of variables $S_e$ is said to *separate* $x_i$ from $x_j$ if all paths between $x_i$ and $x_j$ are *separated* by $S_e$.

(b) A path $P$ is separated by a subset $S_e$ of variables if at least one pair of successive links along $P$ is *blocked* by $S_e$.

**Definition 4.6:**

(a) Two links meeting head-to-tail or tail-to-tail at node $x$ are *blocked* by $S_e$ if $x$ is in $S_e$.

(b) Two links meeting head-to-head at node $x$ are *blocked* by $S_e$ if neither $x$ nor any of its descendants is in $S_e$.

**Theorem 4.3:** Suppose $S$ has the ME distribution subject to the constraint set $CS$. If a set $S_e \subset V$ separates $x_i$ from $x_j$, then $x_i$ is conditionally independent of $x_j$, given $S_e$.

For a proof of this criterion for the general case of belief networks, see [12].

For example, to analyze the social well-being of residents in some mining towns [8], we know that
Alienation at work (A) and/or mental problems (D) may cause some sense of powerlessness (C); and bad living conditions (B) and/or the sense of powerlessness may cause some mental problems.

and in addition to the UCS, we have the following constraint set on the distribution:

$$P(A) = 0.2, \quad P(B) = 0.7,$$
$$P(C|\bar{A},\bar{D}) = 0.1, \; P(C|\bar{A},D) = 0.2,$$
$$P(C|A,\bar{D}) = 0.2, \; P(C|A,D) = 0.6,$$
$$P(D|\bar{B},\bar{C}) = 0.2, \; P(D|\bar{B},C) = 0.3,$$
$$P(D|B,\bar{C}) = 0.4, \; P(D|B,C) = 0.8.$$

Thus we have a belief network in Fig. 4.1 (a) in which there is a directed cycle $C \rightarrow D$, $D \rightarrow C$. The neighbor system is as follows
$\sigma A = \{C,D\}$, $\sigma B = \{C,D\}$, $\sigma C = \{A,B,D\}$, $\sigma D = \{A,B,C\}$.
According to the Markov property of the network, we have
$P(A|B,C,D) = P(A|C,D)$, $P(B|A,C,D) = P(B|C,D)$,
ie. A and B are conditionally independent of each other given C and D. The same result can be obtained by Pearl's graphical criterion (see Fig. 4.1 (b) and (c)). The two paths between A and B are blocked by C and D, respectively.

## 5. Decomposition

Based on Corollary 4.1, we propose the following decomposition method:

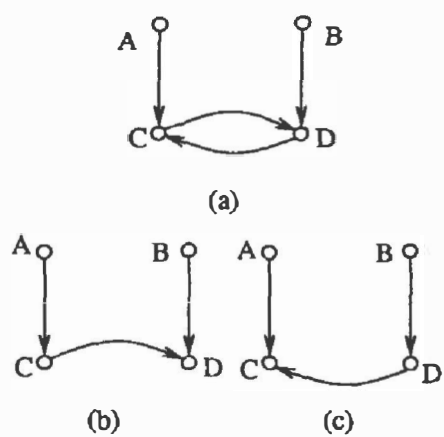

Fig. 4.1

(1) Construct a undirected neighborhood network $G_\sigma = <V, E_\sigma>$ for the belief network $G$, such that $(x_i, x_j) \in E\sigma \leftrightarrow x_i \in \sigma x_j$.

(2) Find a filling-in [10] $F$ of $G_\sigma$, such that there is an $MC$ covering $D_\sigma$ of $G_f = <V, F \cup E_\sigma>$:
$D_\sigma = \{MC_i | MC_i \in MCC \text{ of } G_f, 0 \leq i < n_\sigma\}$
and
$$\sum_{i=0}^{n_\sigma - 1} 2^{|MC_i|} \quad (5.1)$$
has minimum (or sub-minimum) value.

$D_\sigma$ is the decomposition that we want and corresponds to a acyclic hypergraph
$<V, E_h>$, where $E_h = \{MC_i\}$.
Note that the neighborhood network here is equivalent to the moral graph of Spiegelhalter [6]. It can be shown that finding a filling-in with (5.1) minimum is NP hard [14]. Therefore, we propose an algorithm [14] to find an optimal (or suboptimal) filling-in based on simulated annealing [5].

For the example in Fig. 4.1, the neighborhood network is $<V, \{(A,C),(A,D),(B,C),(B,D),(C,D)\}>$. It is already an acyclic hypergraph, thus the filling-in $F = \varnothing$. Its decomposition is $\{\{A,C,D\},\{B,C,D\}\}$. For the network shown in Fig. 5.1, after constructing its neighborhood network the $MCC$
$\{\{A,C,F\},\{B,D,E\},\{C,D\},\{E,F\}\}$
is not an acyclic covering. Thus, we have to look for a filling-in to make the covering acyclic. In this case, we have $F = \{(D,F)\}$, and the acyclic covering is
$\{\{A,C,F\},\{B,D,E\},\{C,D,F\},\{D,E,F\}\}$.

## 6. Consistency of the Constraints

As mentioned earlier, when the network is getting bigger, it becomes very difficult to check the consistency of the constraints. In this section, we propose a local method to check the consistency.

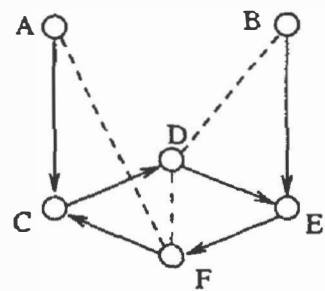

Fig.5.1

According to the theory in acyclic databases [1], we have the following

## Definition 6.1:

(1) A set of relations $r_0, ..., r_{n-1}$ over sets of attributes $R_0, ..., R_n$ are *pairwise consistent* if
$\forall i, j \in \{0, ..., n-1\}, \pi_{R_i \cap R_j}(r_i) = \pi_{R_i \cap R_j}(r_j)$.
where $\pi$ is the projection operator.

(2) $r_0, ..., r_{n-1}$ are *globally consistent* if
$\exists r$ over $\bigcup_{j=0}^{n-1} R_j, \forall i \in \{0, ..., n-1\}, r_i = \pi_{R_i}(r)$.

## Definition 6.2:
We say $R_0, ..., R_{n-1}$ have the running intersection property if the $R_i$'s can be ordered as $S_0, ..., S_{n-1}$ such that
$\forall i \in \{1, ..., n-1\}, \exists j_i < i, ((S_i \cap \bigcup_{j=1}^{i-1} S_j) \subseteq S_{j_i})$. (6.1)

Beeri et al [1] proved

## Theorem 6.1:
The following conditions about $R_0, ..., R_{n-1}$ are equivalent:

(1) $<V, E_h>$ is an acyclic hypergraph, where
$V = \bigcup_{i=0}^{n-1} R_i$, $E_h = \{R_0, ..., R_{n-1}\}$.

(2) Pairwise consistency $\leftrightarrow$ global consistency for $R_0, ..., R_{n-1}$.

(3) $R_0, ..., R_{n-1}$ have the running intersection property.

(4) Graham reduction:

(i) delete attributes that appear in only on $R_i$.

(ii) delete $R_i$ if $\exists R_j$ ($j \neq i, R_i \subseteq R_j$).

reduces $R_0, ..., R_{n-1}$ to nothing if applied repeatedly.

Considering $r$ as the space of all solution vectors of the linear equations set consisting of all constraints and $r_i$'s as the spaces of all solution vectors of the linear equation sets consisting of the constraints on $MC_i$'s which are components of the acyclic decomposition $D_\sigma$, we can see that Theorem 6.1 is enough to support the following consistency checking algorithm:

## Algorithm 6.1:

(1) Order $MC_i$'s in $D_\sigma$ as $S_0, ..., S_{n-1}$ such that (6.1) holds.

(2) Check $S_0$ alone for consistency:
```
if(check_consistency(S_0, constraint[0]) == FALSE) {
    report_inconsistency(S_0, S_0);
    exit(1);
}
```

(3) Check other MC's for pairwise consistency:
```
for(i=1; i<n; i++) {
    C = project(S_{j_i}, constraint[j_i], S_i);
    if(check_consistency(S_i, constraint[i] ∪ C))
                        == FALSE ) {
        report_inconsistency(S_i, S_{j_i});
        exit(1);
    }
}
```

(4) report_consistency();

where C function check_consistency($S_i$, constraint[i]) checks the consistency of constraint set constraint[i] on set $S_i$ according to Theorem 1, and project($S_i$,constraint[i],$S_j$) returns the projection of the constraint set, constraint[i], on the intersection of $S_i$ and $S_j$.

For the example in Fig. 4.1, we can write the conditional constraints $P(C|A, D)$ in the format of linear equations:

$$\begin{cases} P(\bar{A}, \bar{C}, \bar{D}) - 9\, P(\bar{A}, C, \bar{D}) = 0 \\ P(\bar{A}, \bar{C}, D) - 4\, P(\bar{A}, C, D) = 0 \\ P(A, \bar{C}, \bar{D}) - 4\, P(A, C, \bar{D}) = 0 \\ 3\, P(A, \bar{C}, D) - 2\, P(A, C, D) = 0 \end{cases}$$

Its general solution vector is

$(9k_1, 4k_2, k_1, k_2, 4k_3, 2k_4, k_3, 3k_4)$,

where all $k_i$, $1 \leq i \leq 4$ are arbitrary constants. The projection of this solution vector on $\{C, D\}$ is

$$\begin{cases} 9\, P(C, \bar{D}) - P(\bar{C}, \bar{D}) = k'_1 \\ 4\, P(C, D) - P(\bar{C}, D) = k'_2 \end{cases}$$

If the constraint set is consistent then the following sets of linear equations should have at least one non-zero common solution:

$$\begin{cases} P(\bar{B}, \bar{C}, \bar{D}) - 4P(\bar{B}, C, \bar{D}) = 0 \\ 3P(\bar{B}, \bar{C}, D) - 7P(\bar{B}, C, D) = 0 \\ P(B, \bar{C}, \bar{D}) - 3P(B, C, \bar{D}) = 0 \\ 4P(B, \bar{C}, D) - P(B, C, D) = 0 \end{cases}$$

$$\begin{cases} P(\bar{B},\bar{C},\bar{D})+P(\bar{B},\bar{C},D)+P(\bar{B},C,\bar{D})+P(\bar{B},C,D) = 0.3 \\ P(B,\bar{C},\bar{D})+P(B,\bar{C},D)+P(B,C,\bar{D})+P(B,C,D) = 0.7 \end{cases}$$

$$\begin{cases} 9P(\bar{B},C,\bar{D})+9P(B,C,\bar{D})-P(\bar{B},\bar{C},\bar{D})-P(B,\bar{C},\bar{D})=k'_1 \\ 4P(\bar{B},C,D)+4P(B,C,D)-P(\bar{B},\bar{C},D)-P(B,\bar{C},D)=k'_2 \end{cases}$$

It is not very difficult to verify that the determinant of the combination of these sets of equations is not zero, thus it has unique solution vector which depends on the arbitrary constants $k'_1$ and $k'_2$. Therefore the constraint set is consistent.

## 7. Initial distributions

To obtain an initial distribution, we need only to maximize the entropy of the distribution subject to the linear equality constraints. However it is still quite painful to solve non-linear programming problems all the time although we have developed a numerical method to do so (section 3.1 or [11]). We will show in this section that Jeffrey's rule (or partial Jeffrey's rule) and conditional constraint rule (3.2) can be also used alternately to obtain initial distributions if both marginal and conditional constraints are specified on the underlying distribution. The result obtained by this successive updating method is a good approximation to that of real ME method. The convergence of the successive updating can be significantly speeded up by gradient-threshold principle proposed in [13].

Using the numerical method proposed in [11], we can obtain the result for the example in Fig. 4.1 in a few seconds on an IBM-PC. Two marginal for the subspaces $ACD$ and $BCD$ of this result are as follows

| P(A, C, D) | 0.4479 0.2419 0.0498 0.0604 0.0862 0.0369 0.0216 0.0553 |
|---|---|
| P(B, C, D) | 0.1857 0.0464 0.0474 0.0203 0.3484 0.2323 0.0239 0.0954 |

Using above-mentioned successive updating method, we use the constraint with the greatest gradient, $P(A) = 0.2$ ($\nabla(A) = 0.3$), first to update a uniform prior for $ACD$ by Jeffrey's rule and calculate $[P(C,D)]$ (the posterior of the intersection of $ACD$ and $BCD$) from the result, use it to update the uniform prior of $BCD$. Calculating the gradients of all constraints from the posterior in this step. Because the constraint with the greatest gradient is $P(B) = 0.7$ ($\nabla(B) = 0.2$), we use it to do the next step of updating ... After using all the constraints once each, we finish the first cycle of updating and obtain the following distributions for $ACD$ and $BCD$ which is only a coarse approximation to the real ME result.

| P(A, C, D) | 0.4102 0.2397 0.0401 0.0750 0.1086 0.0356 0.0239 0.0668 |
|---|---|
| P(B, C, D) | 0.1694 0.0423 0.0319 0.0137 0.3495 0.2330 0.0320 0.1281 |

The results of the following 4 updating cycles are given in Table 7.1. The last few approximations are quite close (the errors<0.5%) to the real ME result which is obtained without decomposition.

## 8. Conclusions

The problems about how to handle directed cycles in belief networks are discussed in this paper.

(1) It has been shown that general belief networks have quite nice Markov property if the distribution is obtained by the ME/MCE methods.

(2) Based on the above Markov property, the underlying space can be decomposed into small subspaces which are hyperedges in an acyclic hypergraph.

(3) The consistency of the constraints can be checked locally inside each subspace and between two adjacent subspaces according to the running intersection property of the acyclic decomposition.

(4) The initial distribution of the decomposed network is obtained by ME/MCE methods, a successive updating method is recommended which uses Jeffrey's rule and conditional

| Table 7.1 |||
|---|---|---|
| Updating cycle | Subspaces | Distributions |
| 2 | ACD | 0.444003 0.243843 0.042686 0.068039 0.094973 0.032127 0.020544 0.053785 |
| 2 | BCD | 0.200033 0.050008 0.036707 0.015732 0.338943 0.225962 0.026523 0.106092 |
| 3 | ACD | 0.439272 0.247594 0.048808 0.061898 0.091531 0.035206 0.022883 0.052808 |
| 3 | BCD | 0.193179 0.049959 0.042916 0.015804 0.337624 0.232840 0.028775 0.098903 |
| 4 | ACD | 0.446850 0.242888 0.046578 0.063684 0.090022 0.034535 0.021113 0.054330 |
| 4 | BCD | 0.193906 0.048530 0.042647 0.018209 0.342965 0.228894 0.025045 0.099804 |
| 5 | ACD | 0.444434 0.244491 0.049382 0.061123 0.088470 0.035993 0.022118 0.053990 |
| 5 | BCD | 0.190041 0.048267 0.045553 0.018225 0.342863 0.232217 0.025946 0.096887 |

constraint rule alternately and uses gradient-threshold principle to speed up the convergence and to control the termination of the iteration.

(5) The rest of the reasoning can be performed by methods such as those in [6,7,13].

Our experiments show that the successive updating recommended here is about 20 times faster than the conjugate gradient numerical method, and the latter is in turn much faster than some methods like simulated annealing which often need several hundred runs to reach a frozen point. However, to find an optimal or suboptimal filling-in, a method based on simulated annealing may be necessary because the objective function is a multi-model one in contrast to that the entropy function is strictly concave.

The basic idea in this paper has been incorporated into PESS — a Probabilistic Expert System Shell [15] (originally named µ-Shell). A sociologist expert system has been built based on PESS to analyze the social well-being of residents in some mining towns in three states of Australia [8].

## Acknowledgment

Thanks to R. Quinlan, E. A. Sonenberg, B. Marksjo and C. Neil for instructive discussion and encouragement. This author would like also to thank M. Herion, P. Cheeseman, J. Lemmer and two anonymous referees for their comments and criticism.

## REFERENCES

1. C. Beeri and D. Maier, "Properties of Acyclic Database Schemes", in *Proc. 13th Annual ACM Symposium on the Theory of Computing*, ACM, New York, 1981, 355-362.

2. P. Cheeseman, "A Method of Computing Generalized Bayesian Probability Values for Expert Systems", in *Proc. IJCAI 83*, Karlsruhe, W. Germany, 1983, 198-202.

3. P. Diaconis and S. L. Zabell, "Updating Subjective Probability", *Journal of the American Statistical Association*, 77, 380 (1982), 822-830.

4. J. G. Kemeny, J. Snell and A. W. Knapp, in *Denumerable Markov Chains*, Springer, Heidelberg, New York, Berlin, 1976.

5. S. Kirkpatrick, C. D. G. Jr. and M. P. Vecchi, *Optimization by simulated annealing*, IBM Thomas J. Watson Research center, Yorktown Heights, NY., 1982.

6. S. L. Lauritzen and D. J. Spiegelhalter, "Local Computations with Probabilities on Graphical Structures and Their Application to Expert Systems", *J. R. Statist. Soc. B*, 50, 2 (1988).

7. J. F. Lemmer, "Generalized Bayesian Updating of Incompletely Specified Distributions", *Large Scale Systems*, 5 (1983), Elsevier Science Publishers.

8. C. C. Neil, "Regional Centre or Single Enterprise Town: Implications for the Social Well-being of Residents", *Technical Report*, Division of Building Research, Commonwealth Scientific and Industrial Research Organization, Australia, 1988.

9. J. Pearl, "Fusion, propagation, and structuring in belief networks", *Artificial intelligence*, 29 (1986), 241-228, Elseveir Science Publisher B.V. (North-Holland).

10. R. E. Tarjan and M. Yannakakes, "Simple Linear-time Algorithms to Chordality of Graphs, Test Acyclicity of Hypergraphs, And Selectively Reduce Acyclic Hypergraphs", *SIAM J. Compt.*, 13, 3 (Aug. 1984), 566-579.

11. W. X. Wen, "Analytical and Numerical Methods for Minimum Cross Entropy problems", *Technical Report 88/26*, Comput. Science., The University of Melbourne, 1988.

12. W. X. Wen, "Markov and Gibbs fields and ME/MCE Reasoning", *Technical Report (draft)*, Comput. Science., The University of Melbourne, 1989.

13. W. X. Wen, "Reasoning with Multiple Uncertain Evidence", *Technical Report 88/34*, Computer Science, The University of Melbourne, July 1988.

14. W. X. Wen, "MCE Properties of Recursive Causal Networks and Parallel Reasoning", *In preparation, Technical Report*, Comput. Science., The University of Melbourne, 1989.

15. W. X. Wen, "PESS: A Probabilistic Expert System Shell", *Technical Report 89/10*, Comput. Science., The University of Melbourne, 1989.

16. B. P. Wise, "An Experimental Comparison of Uncertain Inference Systems", *Ph.D. Thesis*, Engineering and Public Policy, Carnegie-Mellon University, June 1986.